\documentclass[11pt]{article}
\usepackage[letterpaper,margin=1in]{geometry}
\usepackage{times}
\usepackage[numbers,sort&compress]{natbib}
\usepackage{amsmath,amssymb,amsthm}
\usepackage{booktabs}
\usepackage{graphicx}
\usepackage{microtype}
\usepackage{tabularx}
\usepackage{xcolor}
\usepackage{url}
\usepackage{tikz}
\usepackage{float}
\usepackage[hidelinks]{hyperref}
\hypersetup{pdftitle={SailSearch: Agent-Assisted Discovery in Combinatorial Optimization},
  pdfauthor={Dhriti Verma, Kanthi Sarpatwar, Vinicius Lima, Dzung Phan, Jayant Kalagnanam}}
\usetikzlibrary{arrows.meta,positioning,fit,calc}
\usepackage{placeins}
\usepackage{algorithm}
\usepackage{algpseudocode}

\newcommand{\method}{SailSearch}
\newcommand{\sd}{\texttt{solvediophant}}
\newcommand{\ZZ}{\mathbb{Z}}
\newcommand{\FF}{\mathbb{F}_2}
\newcolumntype{Y}{>{\raggedright\arraybackslash}X}
\newtheorem{proposition}{Proposition}
\definecolor{ssblue}{HTML}{2A78D6}
\definecolor{ssorange}{HTML}{EB6834}
\title{\method{}: Agent-Assisted Discovery\\in Combinatorial Optimization}
\author{Dhriti Verma \quad Kanthi Sarpatwar \quad Vinicius Lima \quad Dzung Phan \quad Jayant Kalagnanam\\[4pt]
IBM Research, Yorktown Heights, NY, USA}
\date{}

\begin{document}
\maketitle

\begin{abstract}
We present \method{}, a human-guided workflow for algorithmic research with
two language-model agents, and report results on Market Split and maximum
independent set (MIS). For Market Split, our randomized algorithm searches
across multiple lattice bases, dividing computation between enumeration
and continued basis reduction. Across ten random seeds per instance, it
found the first reported CPU-only solutions to 18 of 33 previously open QOBLIB
instances, solving 130 to 133 of the 156 benchmark instances per seed
with a 24-hour cutoff on two CPU cores. For MIS, our generator encodes
planted XOR constraints as graphs with certified optima; on seven
instances with 1,752 to 3,996 vertices, the best solutions returned by
the tested solvers in eight-hour runs were 5 to 15 vertices below the
optimum.
\end{abstract}

\section{Introduction}
\label{sec:intro}

Large language models (LLMs) have begun to contribute to algorithm design
and mathematical research, with FunSearch improving bounds on the cap set
problem \citep{romeraparedes2024funsearch} and AlphaEvolve extending this
approach to matrix multiplication and other open mathematical problems
\citep{novikov2025alphaevolve}. Related systems apply the same cycle of generating
and testing candidates to evolve optimization heuristics with executable
feedback \citep{liu2024eoh,ye2024reevo} and to improve reward functions and
prompts \citep{ma2024eureka,yang2024opro}, and later systems add persistent
memory and literature search so that agents can draw on earlier
experiments and external research
\citep{qu2026coral,liu2026skydiscover,liu2025deepevolve}.

However, on a difficult optimization problem, there may not be a fixed
evaluator, or very little useful information from existing evaluators to
inform the next iteration, so algorithm development should depend on
diagnostic tests that explain search behavior as well as on mathematical
understanding. Some research-agent systems explore this experimental work
\citep{lu2024aiscientist,schmidgall2025agentlab,weng2026deepscientist,scientisttwo2026}
autonomously. Here, we use a human-guided workflow with
two agents and apply it to two challenging open-ended problems in
combinatorial optimization: to construct small adversarial MIS benchmarks,
and design a lattice-search algorithm for Market Split.

The two projects contribute classical baselines and testbeds for future
quantum algorithms, motivated by the Quantum Optimization Benchmarking
Library (QOBLIB) \citep{koch2026qoblib}. For MIS, we construct graphs with certified
planted optima that are difficult for the tested general-purpose solvers
to recover: in eight-hour runs on seven graphs with 1,752 to 3,996
vertices, the best results fell short by 5 to 15 vertices.

For Market Split, which asks for a binary vector satisfying a dense
system of integer equations \citep{cornuejols1999marketsplit}, our algorithm
combines a geometric enumeration bound with a schedule that divides the
time budget among searches on multiple lattice bases and continued basis
reduction. Over ten seeds per instance, it found the first reported CPU-only solutions
to 18 of the 33 previously open instances, with 7 to 10 of them solved in
each seed.

We also document how \method{}'s two-agent human-guided workflow reached
these results and discuss how human intervention helped shape the research
process via eleven research episodes tracing the experimental choices that
led to the final generator and solver.

We review related work in Section~\ref{sec:related}, describe the
research setting and \method{} in Sections~\ref{sec:setting}
and~\ref{sec:method}, and present the results of the two projects in
Sections~\ref{sec:mis} and~\ref{sec:marketsplit}. Section~\ref{sec:analysis}
examines how the generator and solver were developed, and the limitations
and conclusion follow in Sections~\ref{sec:limits}
and~\ref{sec:conclusion}. Further results and method details are given in
the appendices, which also list the research episodes.
\section{Related work}
\label{sec:related}

\paragraph{Program search and evaluation feedback.}
Program search has been used for mathematical and algorithmic discovery
in FunSearch and AlphaEvolve
\citep{romeraparedes2024funsearch,novikov2025alphaevolve}, as well as for
optimizing heuristics, prompts, reward functions and machine-learning code
through executable evaluations
\citep{liu2024eoh,ye2024reevo,yang2024opro,ma2024eureka,jiang2025aide}.
Feedback can extend beyond an outcome score, as in ReEvo, which uses
reflection \citep{ye2024reevo}, and GEPA, which refines prompts using
execution traces and textual evaluation feedback \citep{agrawal2026gepa}.
ShinkaEvolve combines novelty filtering with adaptive sampling
\citep{lange2026shinkaevolve}.

\paragraph{Research agents and retained knowledge.}
Research agents now carry out experiments from implementation to written
report
\citep{lu2024aiscientist,yamada2025aiscientistv2,schmidgall2025agentlab,weng2026deepscientist,scientisttwo2026},
while Co-Scientist develops hypotheses for scientists to verify
experimentally \citep{gottweis2026coscientist}. SR-Scientist applies agents
to equation discovery \citep{xia2026srscientist}, and DeepEvolve incorporates
literature research into algorithm evolution \citep{liu2025deepevolve}.
Several systems preserve knowledge from earlier experiments. Arbor
organizes hypotheses and evidence in a tree \citep{jin2026arbor}, and
AutoScientists shares experimental experience among agents
\citep{gao2026autoscientists}. CORAL coordinates asynchronous search
through records of attempts, notes and skills \citep{qu2026coral}.
CliffSearch evolves theory and code together under benchmark feedback
and reviews of correctness and originality \citep{mroueh2026cliffsearch},
while Iris develops diagnostic actions and revises knowledge across
experiments, using a development evaluator to select candidates
\citep{fu2026iris}.
These systems provide precedents for retaining and revising explanations
during algorithm development.

\paragraph{Workflow design and human guidance.}
Automated Design of Agentic Systems and AFlow search over agent
architectures and workflows represented in code, using an archive or
search tree to guide later designs \citep{hu2025adas,zhang2025aflow}.
SkyDiscover provides a modular framework for adaptive search with human
steering through its context interface \citep{liu2026skydiscover}.
The Little Scientist consults a second agent when progress stalls, giving
it a cross-disciplinary prompt without the experiment-history file, and
permits human intervention in the outer research loop. Its revised
preprint reports that a simpler sampling baseline outperformed the full
scaffold in a controlled DNA motif discovery ablation
\citep{smith2026littlescientist}. Eureka accepts human reward initialization
and textual feedback to change a learned policy \citep{ma2024eureka}.
Human control and correction also have a longer history in
mixed-initiative systems and human--AI interaction
\citep{horvitz1999mixedinitiative,amershi2019guidelines}.
In \method{}, human guidance concerns the evidence required for a
decision and the resources committed to testing it, while the ideator agent
examines the worker agent's interpretation from a separate context with access
to the literature.

\paragraph{Studying research decisions.}
LLM-generated ideas received higher novelty ratings than human experts'
ideas in a blinded proposal evaluation \citep{si2025novelideas}, but this
advantage diminished in a subsequent study after experts implemented the
ideas and reviewers assessed the projects \citep{si2026ideationexecution}.
Self-correction without external feedback can also reduce accuracy in
the reasoning settings examined by \citet{huang2024selfcorrect}.
TraceML links successive versions of human and agent code to evaluation
results; its two tested agent systems rarely returned to approaches
they had set aside \citep{yan2026traceml}. Our analysis follows decisions
between proposal and final evaluation, connecting human and agent
interventions to subsequent experiments and algorithmic changes,
including unsuccessful proposals.

\paragraph{Hard instances with known optima.}
Planting a solution in a random constraint system is an established way
to obtain hard satisfiable instances \citep{xu2007rb,jia2007hiding}.
BHOSLIB applies this approach to MIS using model RB: random edges join
disjoint cliques, with one vertex per clique reserved for a planted
independent set and the clique partition certifying the optimum
\citep{xu2004bhoslib}. These graphs have been used as a testbed for vertex-cover local search
\citep{cai2013numvc}, alongside other small benchmarks such as complements
of DIMACS clique instances \citep{johnson1996dimacs} and the
constraint-based graphs and \texttt{evil} family studied by
\citet{sansegundo2023clisat}. \citet{shiraishi2019hardmis} give a
deterministic graph family whose difficulty is established for
cycle-based refutation.

Planted 3-regular 3-XORSAT provides a closer algebraic precedent: these
systems can be solved by Gaussian elimination despite their glassy energy
landscapes \citep{riccitersenghi2010glassy}.
\citet{kowalsky2022xorsat} benchmark heuristic optimizers on random regular
XOR systems encoded as quadratic binary optimization problems with one
auxiliary bit per constraint. Our construction applies girth-preserving
swaps to quasi-cyclic lifts from low-density parity-check (LDPC) code design
\citep{fossorier2004qc,thorpe2003protographs} and encodes each constraint
as a four-vertex clique, with consistency edges that form a path through
each variable's occurrences, following the redundancy principle of reduced
dual CSP encodings
\citep{karakashian2010relational}. Section~\ref{sec:mis-construction}
establishes how this encoding preserves the optimum while admitting
additional partial independent sets. We evaluate them with MIS solvers
that combine data reduction with evolutionary or local search
\citep{lamm2017redumis,dahlum2016onlinemis,grossmann2024mmwis,grossmann2025chils}
and vertex-cover local search methods
\citep{cai2013numvc,cai2015fastvc,luo2019metavc}.

\paragraph{Market Split.}
Cornu\'ejols and Dawande introduced Market Split
\citep{cornuejols1999marketsplit}. Aardal and colleagues
used the Lenstra--Lenstra--Lov\'asz (LLL) basis-reduction algorithm
\citep{lenstra1982lll} to reformulate it as a search in a reduced integer
kernel lattice \citep{aardal2000marketsplit,aardal2000diophantine}, to
which Wassermann applied lattice-point enumeration
\citep{wassermann2002marketsplit}.
Wassermann reports that \sd{} solves every QOBLIB instance with up to
100 variables and every instance with coefficient range 50 up to 130
variables \citep{wassermann2026marketsplit}. Schroeppel and Shamir introduced a subset-sum algorithm with a time--space
tradeoff \citep{schroeppel1981subset}, and Kempke and Koch's GPU variant
reports Market Split results up to 100 variables \citep{kempke2026gpu}.
Wassermann's first-solution experiments run depth-first and
limited-discrepancy enumeration concurrently, and his discussion proposes
copies on different nodes with randomly shuffled input lattice-basis
order \citep{wassermann2026marketsplit}. Our scheduler also allocates
time to continued basis reduction during the search.
\section{Research setting}
\label{sec:setting}

We study two algorithm development projects for combinatorial optimization
problems in which one human researcher worked with agents to generate
adversarial MIS instances and solve Market Split problems. Both projects
took place in a persistent workspace with access to mathematical
literature, public benchmarks, solver implementations, programming tools
and cluster computation, which allowed the agents to implement intermediate
algorithms and diagnostic tests as the research progressed. All logs,
including code and results, were stored between sessions together with the
agents' notes and memory markdown files.

Our analysis follows the development of these two projects through their
preserved records, examining how proposals were tested and how the human
and agents shaped the results.
\section{\method{}}
\label{sec:method}

\begin{figure}[t]
\centering
\begingroup
\definecolor{sailagent}{HTML}{E6EFF7}
\definecolor{sailmemory}{HTML}{F8E9E8}
\definecolor{sailtool}{HTML}{FCF3D9}
\definecolor{sailrule}{HTML}{65717B}
\begin{tikzpicture}[x=1cm,y=1cm,
  font=\normalfont\rmfamily\small,text=black,
  line cap=round,line join=round,
  module/.style={draw=sailrule,line width=0.6pt,rounded corners=2pt},
  flow/.style={-{Latex[length=1.8mm,width=1.25mm]},line width=0.8pt},
  access/.style={draw=black!60,line width=0.65pt},
  symbol/.style={draw=black!85,line width=0.9pt},
  title/.style={font=\normalfont\rmfamily\bfseries\normalsize},
  textlabel/.style={align=center,inner sep=1pt},
  pics/person/.style={code={
    \fill[black!85] (0,0.21) circle (0.115);
    \path[fill=black!85] (-0.26,-0.24) -- (-0.26,-0.15)
      .. controls (-0.26,0.11) and (0.26,0.11) .. (0.26,-0.15)
      -- (0.26,-0.24) -- cycle;
  }},
  pics/terminal/.style={code={
    \draw[symbol,rounded corners=1pt] (-0.39,-0.25) rectangle (0.39,0.28);
    \draw[symbol] (-0.26,0.14) -- (-0.12,0.02) -- (-0.26,-0.10);
    \draw[symbol] (0.03,-0.11) -- (0.23,-0.11);
  }},
  pics/idea/.style={code={
    \draw[symbol] (-0.14,-0.16) -- (-0.14,-0.08)
      .. controls (-0.14,0.01) and (-0.28,0.08) .. (-0.28,0.25)
      .. controls (-0.28,0.59) and (0.28,0.59) .. (0.28,0.25)
      .. controls (0.28,0.08) and (0.14,0.01) .. (0.14,-0.08)
      -- (0.14,-0.16) -- cycle;
    \draw[symbol] (-0.11,-0.25) -- (0.11,-0.25);
    \draw[symbol] (-0.05,-0.32) -- (0.05,-0.32);
    \draw[symbol] (0,-0.16) -- (0,0.11) (-0.10,0.21) -- (0,0.11) -- (0.10,0.21);
  }},
  pics/memory/.style={code={
    \fill[white] (-0.31,-0.22) rectangle (0.31,0.23);
    \draw[symbol] (-0.31,0.23) -- (-0.31,-0.22)
      arc[start angle=180,end angle=360,x radius=0.31,y radius=0.10]
      -- (0.31,0.23);
    \draw[symbol] (0,0.23) ellipse [x radius=0.31,y radius=0.10];
    \draw[symbol] (-0.31,0.01)
      arc[start angle=180,end angle=360,x radius=0.31,y radius=0.10];
  }},
  pics/sailcode/.style={code={
    \draw[symbol] (-0.15,0.23) -- (-0.36,0) -- (-0.15,-0.23);
    \draw[symbol] (0.15,0.23) -- (0.36,0) -- (0.15,-0.23);
    \draw[symbol] (0.07,0.30) -- (-0.07,-0.30);
  }},
  pics/folder/.style={code={
    \draw[symbol,fill=white,rounded corners=1pt]
      (-0.38,-0.25) -- (-0.38,0.27) -- (-0.10,0.27) -- (0.01,0.15)
      -- (0.38,0.15) -- (0.38,-0.25) -- cycle;
  }},
  pics/book/.style={code={
    \draw[symbol] (0,-0.25) -- (0,0.24);
    \draw[symbol] (0,0.24) .. controls (-0.12,0.34) and (-0.26,0.34) .. (-0.38,0.28)
      -- (-0.38,-0.21) .. controls (-0.24,-0.15) and (-0.12,-0.16) .. (0,-0.25);
    \draw[symbol] (0,0.24) .. controls (0.12,0.34) and (0.26,0.34) .. (0.38,0.28)
      -- (0.38,-0.21) .. controls (0.24,-0.15) and (0.12,-0.16) .. (0,-0.25);
  }}]

\pic at (8.2,6.45) {person};
\node[title,textlabel] at (8.2,5.94) {Human researcher};
\node[textlabel] at (8.2,5.52) {Research objectives and process};
\draw[access,densely dotted] (8.2,5.28) -- (8.2,4.98);

\filldraw[module,fill=sailagent] (2.9,3.15) rectangle (5.5,5.15);
\filldraw[module,fill=sailagent] (10.9,3.15) rectangle (13.5,5.15);
\pic at (4.2,4.55) {terminal};
\node[title,textlabel] at (4.2,3.83) {Worker};
\pic[scale=0.85] at (12.2,4.50) {idea};
\node[title,textlabel] at (12.2,3.83) {Ideator};
\draw[flow] (5.5,4.48) -- (10.9,4.48);
\node[textlabel] at (8.2,4.74) {Results and open questions};
\draw[flow] (10.9,3.72) -- (5.5,3.72);
\node[textlabel] at (8.2,3.98) {Proposals and critiques};

\foreach \x/\name in {1.20/Worker,15.20/Ideator}{
  \filldraw[module,fill=sailmemory] (\x-1.10,3.15) rectangle (\x+1.10,5.15);
  \pic at (\x,4.55) {memory};
  \node[textlabel] at (\x,3.70) {\name\\memory};
}
\draw[{Latex[length=1.5mm]}-{Latex[length=1.5mm]},access] (2.30,4.15) -- (2.90,4.15);
\draw[{Latex[length=1.5mm]}-{Latex[length=1.5mm]},access] (13.50,4.15) -- (14.10,4.15);

\draw[access] (4.2,3.15) -- (4.2,2.00);
\draw[access] (12.2,3.15) -- (12.2,2.00);
\draw[access] (5.00,3.15) -- (5.00,2.53) -- (7.35,2.53) -- (7.35,2.00);
\draw[access] (11.40,3.15) -- (11.40,2.53) -- (9.05,2.53) -- (9.05,2.00);
\node[textlabel,fill=white] at (6.22,2.53) {Read / write};
\node[textlabel,fill=white] at (10.20,2.53) {Read};
\foreach \x in {3.10,8.20,13.30}{
  \filldraw[module,fill=sailtool] (\x-1.80,0.10) rectangle (\x+1.80,2.00);
}
\pic at (3.10,1.48) {sailcode};
\node[title,textlabel] at (3.10,0.85) {Sandbox};
\node[textlabel] at (3.10,0.40) {Tests and compute};
\pic at (8.20,1.48) {folder};
\node[title,textlabel] at (8.20,0.85) {Shared repository};
\node[textlabel] at (8.20,0.40) {Code and results};
\pic at (13.30,1.48) {book};
\node[title,textlabel] at (13.30,0.85) {Literature search};
\node[textlabel] at (13.30,0.40) {Papers and references};
\end{tikzpicture}
\endgroup
\caption{The \method{} workflow. The worker studies the problem and runs experiments;
the ideator reads the literature and proposes directions from a separate context.
Both can inspect the experimental repository and keep their own notes. The human researcher passes messages between them and guides
the research objectives and experimental procedure.}
\label{fig:roles}
\end{figure}
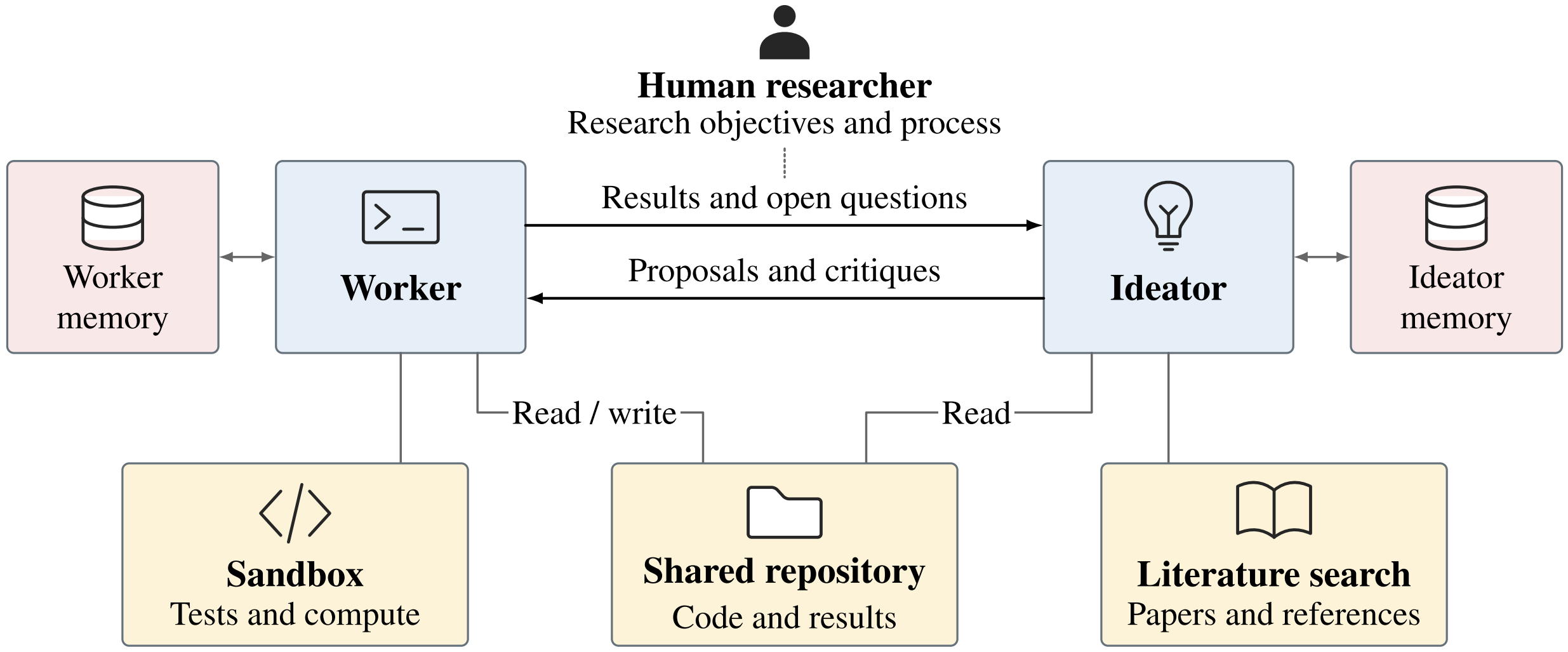

\method{} divides an algorithm development project between two LLM
agents in a sandbox-style environment and a human researcher
(Figure~\ref{fig:roles}).

The worker agent is encouraged to investigate the mathematics of the
challenging problem and to use the compute cluster and its installed
solvers to test intermediate and candidate algorithms in its process of
achieving the user-set goal. In this process, it designs diagnostic tests,
for example of whether a proposed hardness criterion predicts solver
failure or of how search cost varies with the lattice basis. Each result
produced is saved with its code, logs and experimental conditions,
alongside the worker's retained conversation context and notes, so that a
later worker or ideator session can rerun the test or reconsider the
interpretation on which a subsequent decision was based, in a branching
tree of ideas.

Working from a separate conversation history, the ideator consults its
literature notes and the worker's repository to examine assumptions behind
the current approach, and may question an existing interpretation or
propose a construction or experiment, such as the MIS designs drawn from
error-correcting codes
(Section~\ref{sec:trajectories}). The isolation of the two agents and the
interventions by the human help mitigate self-correction-based bias \citep{huang2024selfcorrect}
and context rot \citep{liu2024lostmiddle,hong2025contextrot} associated with
long-running agentic sessions.

Messages between the agents are relayed by the human, who can request
further tests or revise acceptance criteria and compute allocations when
the worker generalizes from a few failed experiments or sets up a test
incorrectly. These interventions are examined in Section~\ref{sec:census}
using the research records.

A cycle begins when progress on the goal stalls or a conclusion needs checking: the
human passes the worker's results and open questions to the ideator, and
the worker turns the reply into executable tests. Depending on the
outcome, the worker may change the problem representation or the
diagnostic used to choose the next experiment, and in Market Split such
outcomes also changed how computation was allocated. For example, after
the ideator proposed searching equivalent lattice bases, a diagnostic
using a known solution found a $2^{12}$-fold difference in search cost
between the best and median bases at 90 variables, which prompted tests of
basis selection using only problem data (Section~\ref{sec:analysis}).

Because earlier implementations are preserved, they can be compared with
later proposals, and a direction can be resumed after the research has
moved on.
In both projects, the worker was
Claude in Claude Code (Opus 4.8, later Opus 5 and 5.5), and the ideator was
GPT-5.5 in ChatGPT. The sessions and repository history, described in
Appendix~\ref{app:corpus}, are analyzed in Section~\ref{sec:analysis}.
\section{Adversarial maximum independent set instances}
\label{sec:mis}

We developed a generator of small adversarial MIS instances with a
planted optimum, motivated by the QOBLIB benchmark. In the generator, a quasi-cyclic system of
XOR constraints is randomized with girth-preserving edge swaps and then
encoded as a sparse MIS graph. Because the planted independent set attains a matching upper bound, we
can measure each solver's shortfall from the optimum exactly on the nine
benchmark instances, which range from 960 to 3,996 vertices.
We report the seven largest instances as a group and also evaluate an
earlier 700-vertex instance and smaller controls.

\subsection{Generating planted instances}
\label{sec:mis-construction}

We start from a 3-regular 3-XORSAT system with \(M\) equations and \(M\)
binary variables: each equation contains three variables, and each
variable appears in three equations. In matrix form, the system is $Hx=c$ over $\FF$, where $H$ is the
binary coefficient matrix, $x$ is the assignment vector, and $c$ specifies
the required parities. We choose a binary assignment $x^*$ and set
$c=Hx^*$.

To generate $H$, we construct a bipartite incidence graph whose two vertex
classes represent constraints and variables, with each edge recording a
one in $H$.
We start from a quasi-cyclic lift of the complete bipartite graph
$K_{3,3}$, a construction used for low-density parity-check (LDPC) codes
\citep{fossorier2004qc,thorpe2003protographs}. Each vertex of $K_{3,3}$ is
replaced by $L$ copies, and each edge by $L$ edges that connect the copies
of its two endpoints in a cyclically shifted order, with the shift chosen
separately for each edge. This gives $M=3L$ constraints and variables, and in $H$ it
produces a $3\times3$ array of $L\times L$ cyclically shifted identity
matrices. We then apply degree- and girth-preserving random edge swaps
\citep{fossorier2004qc,thorpe2003protographs}. The resulting incidence graphs
have girth 12, so every cycle in them has length at least 12 and, in
particular, two constraints share at most one variable.

To build the MIS graph $G$, we create one vertex for each of the four
satisfying assignments to a constraint's three variables and join them
into a clique, so an independent set can select at most one assignment
per constraint. For each variable $x_j$, we randomly order the three cliques corresponding
to constraints that contain it and connect every pair of vertices in
consecutive cliques whose assignments disagree on $x_j$, thereby enforcing
consistency along a path through its three occurrences. In the resulting graph $G$, which
has $4M$ vertices, selected assignments in consecutive cliques must agree
on the shared variable (Algorithm~\ref{alg:mis}).

\begin{algorithm}[htbp]
\caption{Planted MIS instance generator}
\label{alg:mis}
\small
\begin{algorithmic}[1]
\Require lift factor $L\geq1$, swap budget $0\leq s\leq3{,}000$, seed
\Ensure graph $G$, size-$M$ independent set $I^*$, clique partition $\{\mathcal C_i\}_{i=1}^M$
\State initialize the random generator from the seed; $M\gets3L$
\State $H\gets$ a quasi-cyclic lift of $K_{3,3}$ selected by sampled girth
\Statex \hspace{\algorithmicindent} Try at most 40 lifts, stopping at sampled girth at least 12; retain the best.
\State apply up to $s$ degree-preserving edge swaps, creating no cycle shorter than 12
\Statex \hspace{\algorithmicindent} Stop after $20s$ proposals; record $\operatorname{rank}_{\FF}(H)$ after the swaps.
\State draw $x^*\in\FF^M$ uniformly; $c \gets Hx^*$
\For{each constraint $i$}
  \State $\mathcal C_i \gets$ clique on the four assignments of its variables with parity $c_i$
\EndFor
\State $G\gets$ the disjoint union of $\mathcal C_1,\ldots,\mathcal C_M$
\For{each variable $x_j$}
  \State $(i_1,i_2,i_3) \gets$ random order of the constraints containing $x_j$
  \State join $u\in \mathcal C_{i_1}, v\in \mathcal C_{i_2}$ and $u\in \mathcal C_{i_2}, v\in \mathcal C_{i_3}$ whenever $u,v$ disagree on $x_j$
\EndFor
\State $I^* \gets$ the set containing the vertex of each $\mathcal C_i$ that agrees with $x^*$
\State apply one random vertex permutation to $G$, $I^*$ and every $\mathcal C_i$
\State \Return $G$, $I^*$, $\{\mathcal C_i\}_{i=1}^M$ \Comment{the set and partition certify optimality}
\end{algorithmic}
\end{algorithm}

\begin{samepage}
\begin{proposition}
\label{prop:mis}
The maximum independent sets of $G$ have size $M$ and correspond one to one
to solutions of $Hx=c$. If $H$ has full rank over $\FF$, the maximum
independent set is unique.
\end{proposition}

\begin{proof}
The clique partition bounds the independent set size by $M$. Every solution
of $Hx=c$, including $x^*$, attains this bound by selecting its local
assignment in each clique. Conversely, an independent set of size $M$
selects one assignment per constraint. The path edges make all three
occurrences of each variable agree, so these local assignments define a
solution of $Hx=c$. Full rank makes that solution unique.
\end{proof}
\end{samepage}

We checked the rank after the edge swaps: all nine benchmark matrices have
full rank, so their planted optima are unique.

Our path encoding connects two clique pairs per variable, compared with
three in the all-pairs encoding. Since each value of the shared variable occurs in two local assignments
per clique, each paired set of cliques contributes eight conflict edges,
two incident to each vertex. Including the six edges within each clique, the path encoding has
$22M$ edges, compared with $30M$ for all-pairs consistency. All four vertices
within a clique have equal degree.

For each variable, omitting the edges between the first and third cliques
admits partial independent sets that leave the middle clique empty and
select assignments at the two ends that disagree on that variable.
Every independent set of the all-pairs encoding remains valid. At size $M$, every clique is occupied, so agreement propagates along
each path. Hence, both encodings have exactly the same maximum independent
sets. Further generation
and certificate details are in Appendix~\ref{app:mis-method}.

\subsection{Solver performance}
\label{sec:mis-solvers}

We evaluated the instances with thirteen solver configurations, ranging
from MIS and vertex-cover local search to memetic and integer-programming
methods:
ReduMIS, OnlineMIS, mmwis, mmwiss, CHILS, NuMVC, two FastVC implementations,
three MetaVC configurations, CP-SAT and CPLEX
\citep{lamm2017redumis,dahlum2016onlinemis,grossmann2024mmwis,grossmann2025chils,cai2013numvc,cai2015fastvc,luo2019metavc}.
On the nine graphs, 154 eight-hour runs were performed, with new
evaluation seeds for the randomized heuristics. The run counts and the
results of each configuration are reported in
Appendix~\ref{app:mis-results}.

\begin{table}[tb]
\centering
\caption{MIS benchmark results. Best is the largest independent set found over
the eight-hour runs; shortfall is the planted optimum minus Best.}
\label{tab:mis}
\small
\begin{tabular}{rrrrrr}
\toprule
Vertices & Edges & Optimum & Runs & Best & Shortfall\\
\midrule
960 & 5{,}280 & 240 & 24 & 239 & 1 \\
1{,}356 & 7{,}458 & 339 & 20 & 336 & 3 \\
1{,}752 & 9{,}636 & 438 & 20 & 433 & 5 \\
2{,}160 & 11{,}880 & 540 & 15 & 533 & 7 \\
2{,}556 & 14{,}058 & 639 & 15 & 630 & 9 \\
2{,}952 & 16{,}236 & 738 & 15 & 729 & 9 \\
3{,}360 & 18{,}480 & 840 & 15 & 828 & 12 \\
3{,}756 & 20{,}658 & 939 & 15 & 924 & 15 \\
3{,}996 & 21{,}978 & 999 & 15 & 984 & 15 \\
\bottomrule
\end{tabular}
\end{table}

All 154 runs finished below the planted optimum (Table~\ref{tab:mis}),
with the best shortfalls on the seven largest instances ranging from
5 to 15 vertices.
NuMVC and the MetaVC configurations achieved the smallest mean relative
shortfalls, below 1.5\% on these instances. In an earlier evaluation with two-hour runs, no optimum was found either,
and increasing the budget from 2 to 8 hours improved the best observed result
on only three of the nine graphs, by at most two vertices. In a separate study, we compared incidence
designs and consistency encodings on newly generated instances at a
fixed ten-minute cutoff (Appendix~\ref{app:mis-components}).

\subsection{Structural analysis}
\label{sec:mis-analysis}
\label{sec:mis-frontier}
\label{sec:mis-limits}

Using the underlying linear system, we can calculate how far a nearly
satisfying assignment is from the planted optimum: an assignment violating
precisely the constraints in $S$ differs from $x^*$ by $H^{-1}\mathbf 1_S$, where
$\mathbf 1_S$ is the indicator vector of $S$ over $\FF$. Among assignments violating
two constraints, the minimum Hamming distance to $x^*$ ranges from 86 to
404 variables across the nine graphs, approximately in proportion to
graph size. Thus, even an assignment violating only two constraints can be far from
the optimum in Hamming distance.

Because partial independent sets may contain inconsistent variable
assignments, we compared the independent sets returned by the solvers using only
variables that receive the same value from every selected vertex
containing them, finding disagreement with
$x^*$ on roughly half of these consistently decoded positions
(Appendix~\ref{app:mis-method}).

However, these instances can be solved quickly by a decoder that
recognizes the XOR structure: it recovers the constraint cliques and
shared variables from the graph, then solves the reconstructed system by
Gaussian elimination. An optimum was returned for each of the nine
instances in under 0.08 seconds, which makes these graphs a testbed for comparing general-purpose
search with methods that exploit the encoded algebraic structure.
\section{Market Split}
\label{sec:marketsplit}

\paragraph{Problem.} A Market Split instance consists of $m$ integer equations
in $n$ binary variables. Given a coefficient matrix $A\in\ZZ^{m\times n}$
and a target vector $b\in\ZZ^m$, the task is to find $x\in\{0,1\}^n$
satisfying $Ax=b$.

In QOBLIB, $n=10(m-1)$ and the entries of $A$ lie in
$\{0,\dots,D-1\}$.
Each target is half the corresponding row sum, rounded down, so
$b=\lfloor A\mathbf 1/2\rfloor$. With four instances for each class
$(m,n,D)$, where $m=3,\dots,15$ and $D\in\{50,100,200\}$, the benchmark
contains 156 instances.
Cornu\'ejols and Dawande proposed instances of this form as a challenge
for integer programming \citep{cornuejols1999marketsplit}, and even small
instances proved very hard for branch and bound based on linear
programming \citep{aardal2000marketsplit}. Because $b=\lfloor
A\mathbf 1/2\rfloor$, the center of the binary box, $\tfrac12\mathbf 1$,
satisfies every equation to within one half, so the linear-programming
relaxation is feasible and gives the search little guidance about which
variables to fix.

We call an instance \emph{previously solved} if the QOBLIB leaderboard marks it
solved \citep{qoblibleaderboard} or \citet{wassermann2026marketsplit} reports a
first-solution time for all four instances of its class.

\subsection{Lattice-search algorithm}

Using the integer-kernel reformulation developed for Market Split
\citep{aardal2000marketsplit,aardal2000diophantine}, we search for binary
points in lattice coordinates under a geometric bound, dividing effort
between enumeration on multiple bases and continued reduction of those
bases as the search proceeds.

For a variable ordering $\pi$, let $A_\pi$ be $A$ with its columns in that
order. Following Aardal et al., a lattice basis that embeds $A_\pi$ and
$b$ with a large scaling factor is reduced with LLL or with its block
generalization, block Korkine--Zolotarev (BKZ) reduction
\citep{schnorr1994bkz}, which finds shorter vectors at a cost that grows
with the block size. The reduced basis yields a particular integer
solution $x_d$ with $A_\pi x_d=b$ and a matrix $K$ whose rows form a basis
of the full integer kernel, satisfying $A_\pi K^{\mathsf T}=0$
\citep{aardal2000diophantine,wassermann2026marketsplit}. Since the QOBLIB matrices have full row rank, $K$ has $n-m$ rows, and any
integer combination of them can be added to $x_d$ while preserving the
equations. Therefore, we search over the integer coefficients $e$ in
\[
  z(e)=x_d+K^{\mathsf T}e,\qquad e\in\ZZ^{n-m},
\]
seeking a vector $z(e)$ whose entries are all zero or one. A Market Split solution is then recovered by restoring the original
variable order.

\paragraph{Geometric bound.}
The distance from the same box center, $\tfrac12\mathbf 1$, also
distinguishes binary vectors from all other integer vectors. Indeed,
for an integer vector $z$,
\[
  \|z-\tfrac12\mathbf 1\|^2
  =\frac n4+\sum_{j=1}^n z_j(z_j-1).
\]
Every summand vanishes when $z$ is binary, giving squared distance
$n/4$, whereas any nonbinary integer coordinate contributes at least two,
so every other integer vector has squared distance at least $n/4+2$.

To apply this gap during enumeration, we factor $K^{\mathsf T}=QR$,
where $Q$ has orthonormal columns and $R$ is upper triangular, and project
the displacement $\delta=\tfrac12\mathbf 1-x_d$ from $x_d$ to the box center
onto the basis span to obtain $y=Q^{\mathsf T}\delta$. The squared
perpendicular distance, $C_\perp=\|\delta\|^2-\|y\|^2$, contributes the same
amount to the distance of every lattice point from the box center. Thus
\[
  \|z(e)-\tfrac12\mathbf 1\|^2=C_\perp+\|Re-y\|^2.
\]
We place a squared-distance threshold midway across the gap and subtract
this fixed contribution, which gives the enumeration bound
\begin{equation}
  \|Re-y\|^2 < E_{\max},\qquad E_{\max}=\frac n4-C_\perp+1.
  \label{eq:ms-geometric}
\end{equation}
Hence, in exact arithmetic, an integer coefficient vector satisfies this
bound if and only if $z(e)$ is binary. Since $R$ is upper triangular, the coefficients can be chosen from
$e_{n-m}$ back to $e_1$, and each choice fixes one additional squared
residual, whose addition to the partial sum gives a lower bound on $\|Re-y\|^2$ and permits pruning when that sum exceeds
$E_{\max}$.

\paragraph{Cost-bounded enumeration.}
For each coefficient $e_k$, given the values already chosen for
$e_{k+1},\ldots,e_{n-m}$, enumeration visits integers in order of distance
from their conditional center, following the nearest-first ordering of
Schnorr--Euchner enumeration \citep{schnorr1994bkz}. Within the geometric
bound, we guide the search with two path costs. The rank cost depends on each
integer's position in this order and on the corresponding diagonal entry
of $R$. The Gaussian cost uses discrete-Gaussian weights
centered at the conditional center, adapting the conditional
probabilities in Klein's randomized nearest-plane method
\citep{klein2000}. A separate depth-first traversal is run for each path cost on each basis,
with an
adaptive threshold that bounds the accumulated path cost and prunes any
branch whose cost exceeds it. Related approaches to
controlling lattice enumeration include extreme pruning
\citep{gama2010extreme} and discrete pruning \citep{aono2017discrete}.

\paragraph{Allocating effort among bases.}
Because the enumeration needed to find a solution depends on the basis,
we distribute the search over two single-threaded processes, each managing
up to 12 bases. The bases differ in column ordering and reduction
choices, including integer row additions followed by BKZ reduction, and
within each process they are searched sequentially.

\begin{algorithm}[!htb]
\caption{Lattice search over multiple bases (one of two processes)}
\label{alg:ms}
\small
\begin{algorithmic}[1]
\Require $A\in\ZZ^{m\times n}$, $b\in\ZZ^m$, seed, process index (0 or 1), time budget $B$
\Ensure a verified solution if found; otherwise no solution found within the budget
\State assign the 12 of the 24 seed-generated configurations whose index parity equals the process index
\State initialize an empty search list $\mathcal F$ and a queue of bases awaiting continued reduction; $U\gets0$
\State build configurations until a basis is available, or configurations or time run out
\For{round $r=0,1,\ldots$ while elapsed time $\tau<B$ and $\mathcal F\ne\emptyset$}
  \State build further configurations up to $\min(2^r,12)$ total attempts, as time permits
  \For{each search $F$ in a snapshot of $\mathcal F$}
    \While{a basis is queued, $\tau<B$ and $U<(\tau-600)/2$}
      \State continue reducing the oldest queued basis with BKZ-50, BKZ-60 and BKZ-70; add the time to $U$
      \State on success, replace its two searches with newly initialized searches
    \EndWhile
    \State \textbf{if} $F$ was replaced \textbf{then continue}
    \State $q\gets0$ \Comment{nodes used by $F$ in this round}
    \While{$q<2^{16+r}$ and $\tau<B$}
      \State resume $F$ for at most $2^{16+r}-q$ nodes, pruning by geometry and path cost
      \State add the nodes used to $q$ and to $F$'s current-traversal count
      \If{a leaf passes the geometric bound}
        \State reconstruct $x=x_d+K^{\mathsf T}e$ in the original variable order
        \State \textbf{if} $x\in\{0,1\}^n$ and $Ax=b$ \textbf{then return} $x$
        \State \textbf{continue} \Comment{resume after a rejected leaf}
      \EndIf
      \State \textbf{if} the traversal is unfinished \textbf{then break}
      \State record its threshold and node count; increase the threshold and restart
    \EndWhile
  \EndFor
\EndFor
\State \Return no solution found by this process
\end{algorithmic}
\end{algorithm}

Starting from one reduced basis per process, we double the target number
of active bases in successive rounds, up to 12, as well as the node budget
for each search (Algorithm~\ref{alg:ms}). A search that reaches its node budget is resumed from the same position
in a later round, and a completed traversal is restarted with a higher
path-cost threshold, whose increment is chosen from previous node counts.
As a result, the schedule adds new bases while allowing larger searches on
those already available.

After an initial search period, continued reduction, BKZ with larger
block sizes, is applied to existing bases whenever the time spent on it is
below a prescribed share of elapsed time, and each newly reduced basis
replaces the earlier one and starts two new traversals. The path costs
and schedule parameters are specified in Appendix~\ref{app:ms-method}.

\subsection{Evaluation}

We ran the algorithm with ten different random seeds on each of the 156
instances, using two
single-threaded processes and two CPU cores per run, with a 24-hour
cutoff. We compare against CPLEX 22.1.2, \sd{} v1.0.0 and the CPU--GPU
algorithm of \citet{kempke2026gpu}, each evaluated at the same cutoff.
For all four algorithms, we used Intel Xeon Platinum 8468 processors,
reserving 4\,GiB of memory for each lattice run and setting a 16\,GiB
memory limit for CPLEX and \sd{}, each of which used two CPU cores. The GPU runs
used 32 CPU cores, host-memory allocations of 256--480\,GiB and one
NVIDIA H200 with approximately 140\,GiB of device memory.

For \sd{}, we followed Wassermann's first-solution protocol: concurrent
depth-first search (DFS) and limited-discrepancy search (LDS), with BKZ
block size 40 and LDS discrepancy limit 20
\citep{wassermann2026marketsplit}. The GPU algorithm was run with the authors'
unmodified implementation.
All the reported runtimes are wall-clock times.

\begin{table}[tb]
\centering
\caption{Market Split classes with at least 100 variables, four instances per
class. Each method is evaluated at a 24-hour cutoff per run. Columns give the number
solved (of 4) and the successful runtime mean $\pm$ sample SD in hours.
For the lattice algorithm, solved counts are ranges over ten seeds and
times summarize all successful runs. Each baseline has one run per
instance. A dash means no success; NA means only one successful run.
Values below 0.01 hours are shown as $<0.01$.
$\dagger$: memory exhausted on at least one instance
(CPLEX: 16\,GiB host; Kempke--Koch: 480\,GiB host or approximately
140\,GiB device memory).
}
\label{tab:frontier}
\footnotesize
\setlength{\tabcolsep}{3pt}
\begin{tabular}{rcrcrcrcr}
\toprule
& \multicolumn{2}{c}{CPLEX} & \multicolumn{2}{c}{Kempke--Koch} & \multicolumn{2}{c}{\sd{}} & \multicolumn{2}{c}{Lattice algorithm}\\
\cmidrule(lr){2-3}\cmidrule(lr){4-5}\cmidrule(lr){6-7}\cmidrule(lr){8-9}
$n/D$ & Solved & Mean $\pm$ SD & Solved & Mean $\pm$ SD & Solved & Mean $\pm$ SD & Solved & Mean $\pm$ SD\\
\midrule
100/50 & 0$^{\dagger}$ & --- & 4 & $8.57\pm 2.03$ & 4 & $<0.01\pm <0.01$ & 4 & $<0.01\pm <0.01$ \\
100/100 & 0$^{\dagger}$ & --- & 0 & --- & 2 & $3.39\pm 1.43$ & 4 & $0.31\pm 0.35$ \\
100/200 & 0$^{\dagger}$ & --- & 0 & --- & 1 & $1.68\pm \mathrm{NA}$ & 4 & $0.88\pm 0.91$ \\
110/50 & 0$^{\dagger}$ & --- & 0$^{\dagger}$ & --- & 4 & $0.19\pm 0.13$ & 4 & $0.03\pm 0.03$ \\
110/100 & 0$^{\dagger}$ & --- & 0$^{\dagger}$ & --- & 2 & $23.11\pm 1.22$ & 4 & $3.67\pm 3.02$ \\
110/200 & 0$^{\dagger}$ & --- & 0$^{\dagger}$ & --- & 0 & --- & 2--4 & $5.85\pm 4.87$ \\
120/50 & 0$^{\dagger}$ & --- & 0$^{\dagger}$ & --- & 4 & $1.63\pm 1.92$ & 4 & $0.43\pm 0.47$ \\
120/100 & 0$^{\dagger}$ & --- & 0$^{\dagger}$ & --- & 0 & --- & 0--3 & $14.03\pm 6.10$ \\
120/200 & 0$^{\dagger}$ & --- & 0$^{\dagger}$ & --- & 0 & --- & 0--1 & $11.16\pm 6.70$ \\
130/50 & 0$^{\dagger}$ & --- & 0$^{\dagger}$ & --- & 2 & $10.27\pm 12.21$ & 4 & $3.67\pm 1.91$ \\
130/100 & 0$^{\dagger}$ & --- & 0$^{\dagger}$ & --- & 0 & --- & 0--1 & $22.85\pm \mathrm{NA}$ \\
130/200 & 0$^{\dagger}$ & --- & 0$^{\dagger}$ & --- & 0 & --- & 0--1 & $22.22\pm \mathrm{NA}$ \\
140/50 & 0$^{\dagger}$ & --- & 0$^{\dagger}$ & --- & 0 & --- & 1--3 & $11.16\pm 7.27$ \\
140/100 & 0$^{\dagger}$ & --- & 0$^{\dagger}$ & --- & 0 & --- & 0 & --- \\
140/200 & 0$^{\dagger}$ & --- & 0$^{\dagger}$ & --- & 0 & --- & 0 & --- \\
\bottomrule
\end{tabular}
\end{table}

\subsection{Results and component comparisons}

On the full 156-instance suite, our lattice algorithm solved 130--133 instances per seed,
compared with 115 for \sd{}, 93 for the GPU algorithm and 40 for CPLEX
(Table~\ref{tab:frontier}). Over ten seeds, it solved a mean of 131.2
instances, with a sample standard deviation of 1.40. Across all ten seeds,
we found the first reported CPU-only solutions to 18 of the 33 previously open
instances, with 7 to 10 of the 33 solved per seed. Results for all 39
classes, together with per-instance success rates and runtimes for the
first reported CPU-only solutions, are reported in Appendix~\ref{app:ms-results}.

We also evaluated four variants of the final algorithm on all 156 instances:
one basis per process, no continued reduction, rank cost only and
Gaussian cost only. The same five seeds were used for each variant and for the full
algorithm, with two CPU cores and a 24-hour cutoff per run, and all
remaining search
parameters, including the initial basis generation and the geometric
bound, were held fixed.

\begin{samepage}
The full algorithm solved a mean of 131.6 instances per seed, compared
with 128.6 for one basis per process, 127.2 without continued reduction,
and 129.6 and 128.2 for rank cost only and Gaussian cost only,
respectively (Table~\ref{tab:ms-components}). Continued reduction made the
largest observed contribution, and removing it reduced the mean by 4.4
instances. The complete comparisons and uncertainty estimates are reported
in Appendix~\ref{app:ms-components}.
\par
\end{samepage}
\section{How the discoveries were made}
\label{sec:analysis}

From the agent sessions and repository history of seven research phases,
we selected eleven episodes in which a proposal or working explanation
led to an implemented test and a subsequent decision, including proposals
that were rejected (Appendix~\ref{app:episodes}). We examine how these
decisions changed the diagnostics and the allocation of computation that
led to the final generator and solver.

\subsection{Reopening a stalled search}
\label{sec:trajectories}

When the worker treated small unsuccessful experiments as a reason to
stop exploring a larger family of approaches, the human challenged the
conclusion and brought the evidence to the ideator for new directions
and changes to existing ones. In MIS, the worker had assumed a lower size limit for resistant
instances and described random regular XOR systems as the best available
structure. Drawing on error-correcting code design, the ideator proposed six
approaches, including quasi-cyclic lifts of a small base graph, which the worker
implemented and later developed into the final instance family.

When the worker again recommended abandoning approaches after small
tests, the ideator shifted from proposing constructions to questioning
what the results established, asking it to distinguish the configurations
actually tested from conclusions about a larger class of ideas, as the
human had done earlier.

Similarly, in Market Split, the ideator questioned the assumption that
the reduced lattice basis should stay fixed and proposed searching equivalent bases.
Although the worker recommended stopping this line of research after
several early comparisons, further exchanges with the ideator led it to
continue developing enumeration across bases, producing an intermediate
solver that solved three of four development instances with 90 variables
and $D=200$.

\subsection{Testing proxy measures}

Early MIS instances were solved quickly, making the tests inexpensive
but offering little guidance about which structural change to try next,
whereas a Market Split test could take hours and return no solution,
leaving the worker uncertain which bases deserved more computation.
Choosing a proxy measure of progress was therefore part of the research
problem.

Comparisons using a known solution showed that the choice of basis could
greatly change the search needed to reach it, leading the worker to seek
a criterion for choosing promising bases before committing a large
budget. An early rule that favored bases whose candidate solutions had the smallest
recorded violation concentrated more than five million checks of candidate solutions
on one basis without finding a solution, after which the worker, with
input from the ideator, changed to more even allocation across bases.
Nevertheless, search over multiple bases was retained, and only its
allocation rule was revised.

\subsection{Retaining useful alternatives}

Although the worker moved on to other proposals after the first MIS
exchange, the quasi-cyclic generator was kept in the repository with its code and
measurements, so it could be compared with later generators
when the human asked which generator had the strongest evidence.
The worker then returned to this generator, scaling it and sparsifying
its consistency edges to produce the final benchmark family.

\subsection{Human guidance}
\label{sec:census}

Most human corrections concerned objectives, evidence requirements and
computation. For MIS, the requirement that instances resist both CPLEX
and local search ruled out a generator variant after CPLEX solved a test
instance within seconds. In Market Split, the human's question about whether stronger basis
reduction justified its wall-clock cost led the worker to compare
reduction and enumeration time and then develop a schedule that begins
with inexpensive searches and adds continued reduction as those searches
proceed.

These interventions changed the questions the experiments had to answer.
Alongside the corrections to premature conclusions discussed above,
they determined which solvers an instance had to resist and which costs
the search schedule had to account for. The smaller number of technical suggestions and design constraints is
also recorded in the message inventory
(Appendix~\ref{app:human-inventory}).
\section{Limitations}
\label{sec:limits}

Because this retrospective study follows two projects with one human
researcher, it does not establish the causal effects of human guidance, the
second agent or persistent memory, which would require comparisons with
matched models, tools and research budgets.
The earlier MIS investigations used changing size targets and solver
settings, with repeated human and ideator interventions, and the final
instances were evaluated against a substantial but limited set of
general-purpose solvers. Finally, for Market Split, we developed and evaluated the algorithm on
the same QOBLIB suite, so performance on newly generated instances remains
to be measured.
\section{Conclusion}
\label{sec:conclusion}

We presented an MIS instance generator and a randomized lattice-search
algorithm for Market Split, developed using \method{}. The generator
produces graphs with certified unique optima that the tested general-purpose
solvers failed to recover within eight hours, while the lattice algorithm
found, over ten seeds, the first reported CPU-only solutions to 18 of the 33
previously open QOBLIB instances.
The accompanying research records document the experimental choices that
led to both methods. These results extend the available Market Split
baselines and provide MIS benchmarks for studying the gap between
general-purpose search and recovery of the encoded algebraic structure.
\section*{AI use statement}

In addition to the discovery sessions described in
Appendix~\ref{app:corpus}, separate agent sessions assisted with checking
results and references and with drafting the manuscript. The human authors
defined the research questions and acceptance criteria and are responsible
for the manuscript.

\bibliographystyle{plainnat}
\bibliography{references}

\clearpage
\appendix

\section{Maximum independent set: additional results}
\label{app:mis-results}

\subsection{Solver configurations and detailed results}

The evaluation includes thirteen configurations: ReduMIS, OnlineMIS,
mmwis and mmwiss from KaMIS; CHILS; NuMVC; FastVC from LibMVC and Cai's
implementation; MetaVC in its default, DIMACS and BHOSLIB configurations;
CP-SAT; and CPLEX 22.1.2 on the edge formulation. ReduMIS used
its full configuration in the standard graph-partitioning mode, which
selects the more time-consuming parameters, with a reduction threshold of
500,000 unsuccessful iterations. When mmwis or mmwiss stopped under its internal convergence
rule, it was restarted with a new seed until the eight-hour budget was
exhausted, retaining the best independent set found. Patches handled input,
seeds, time limits and output; one FastVC implementation also had an early
stagnation stop removed. Randomized heuristics used new evaluation seeds;
CPLEX used seed 1. Heuristics that accept a stopping target received the
certified optimum. CPLEX received neither an optimum bound nor a planted
assignment.

The eight-hour cluster runs used Intel Xeon Platinum 8468 processors.
Single-threaded solvers received one core and 6\,GiB of memory; CHILS used
eight threads, eight cores and 16\,GiB; CP-SAT used sixteen search workers,
sixteen cores and 32\,GiB. CPLEX ran on an Apple M4 Max with five threads,
a 4,096\,MB working-memory threshold and compressed node files on disk.
The two-hour CPLEX evaluation used six threads.

The eight-hour study contains 24 runs at 960 vertices, 20 each at 1,356
and 1,752 vertices, and 15 on each larger graph. All thirteen
configurations are represented on every graph, but repetitions are
unevenly distributed. The best observed results in
Table~\ref{tab:mis-solvers} therefore depend partly on the number of runs
per configuration. The two-hour comparison contains 285 runs across the
same thirteen configurations.

\begin{table}[htbp]
\centering
\caption{Best shortfall from the planted optimum over the eight-hour runs of each
solver configuration on the seven largest instances. The final column
averages the relative shortfall over the seven instances.}
\label{tab:mis-solvers}
\small
\setlength{\tabcolsep}{4.5pt}
\begin{tabular}{lrrrrrrrr}
\toprule
& \multicolumn{7}{c}{Vertices} & \\
\cmidrule(lr){2-8}
Solver & 1,752 & 2,160 & 2,556 & 2,952 & 3,360 & 3,756 & 3,996 & Mean shortfall (\%)\\
\midrule
MetaVC (DIMACS) & 5 & 7 & 9 & 11 & 12 & 15 & 16 & 1.4 \\
NuMVC & 5 & 7 & 9 & 11 & 13 & 15 & 15 & 1.4 \\
MetaVC (BHOSLIB) & 5 & 8 & 9 & 9 & 13 & 15 & 17 & 1.4 \\
MetaVC (default) & 6 & 7 & 9 & 11 & 13 & 15 & 16 & 1.5 \\
ReduMIS & 6 & 8 & 10 & 12 & 13 & 15 & 16 & 1.5 \\
CHILS & 6 & 8 & 11 & 12 & 15 & 15 & 17 & 1.6 \\
FastVC (LibMVC) & 6 & 9 & 12 & 14 & 16 & 18 & 19 & 1.8 \\
FastVC (Cai) & 7 & 9 & 12 & 14 & 15 & 18 & 20 & 1.8 \\
CP-SAT & 9 & 10 & 13 & 14 & 16 & 18 & 19 & 1.9 \\
OnlineMIS & 12 & 12 & 14 & 16 & 19 & 21 & 26 & 2.3 \\
mmwis & 10 & 16 & 20 & 23 & 31 & 34 & 36 & 3.2 \\
mmwiss & 12 & 16 & 21 & 27 & 32 & 35 & 39 & 3.4 \\
CPLEX & 28 & 56 & 31 & 72 & 48 & 52 & 101 & 7.5 \\
\bottomrule
\end{tabular}
\end{table}

NuMVC and MetaVC had the smallest average relative shortfalls. CPLEX's
best solutions were 28--101 vertices below the optimum, although its
reported upper bounds matched the certified optima to numerical precision.
ReduMIS reported unchanged kernels and zero offset on all nine graphs.

\subsection{Screening and smaller instances}

At 960 vertices, five candidates were screened with NuMVC and MetaVC's
DIMACS configuration, using four seeds per solver and a 30-minute limit.
All 40 runs finished two or three vertices below the optimum. One of these
candidates was retained for the longer evaluation in Table~\ref{tab:mis}.

For smaller graphs from this family, 19 of 22 runs at 156 vertices and
6 of 22 runs at 552 vertices reached the optimum within 30 minutes.
At 648 and 708 vertices, respectively 4 and 3 of 49 confirmation runs
reached the optimum within two hours.

On an earlier 700-vertex instance, the certified optimum is 175. All three eight-hour ReduMIS runs returned 174 using the
configuration above. CP-SAT with sixteen threads found and proved the
optimum in 7.9 seconds. CliSAT with color ordering, WeGotYouCovered and
CPLEX also found and proved the optimum within eight hours.

\subsection{Effects of incidence design and consistency encoding}
\label{app:mis-components}

We generated three independent full-rank XOR systems for each combination
of graph size (960 and 1,752 vertices) and incidence design. The quasi-cyclic design uses
the generator in Appendix~\ref{app:mis-method}, with Tanner girth at least
12. The comparison design samples random 3-regular incidence graphs,
conditioned on distinct variables within each constraint and full rank.
All accepted random-regular systems had girth 4. This comparison therefore
tests the incidence designs as a whole, including their different girths.
Acceptance depended only on these structural conditions and preceded
solver evaluation. Generating the six accepted systems required 20
attempts for the quasi-cyclic design and 42 for the random-regular design.

Each system produces paired graphs with path and all-pairs consistency
edges, sharing the planted assignment, local satisfying assignments and
vertex labels. We evaluated all 24 graphs with NuMVC and default MetaVC,
using three seeds per solver, one core and a ten-minute limit per run.
These 144 runs are separate from the eight-hour benchmark. We first
averaged shortfalls over the three solver seeds for each graph, then
computed the mean and sample standard deviation across the three
independently generated systems.

\begin{table}[htbp]
\centering
\caption{Shortfall from the planted optimum after ten minutes: mean
$\pm$ sample standard deviation across three systems, after averaging
over three solver seeds per graph. Path and all-pairs graphs share the
same XOR system. All 144 runs stopped short of the optimum.}
\label{tab:mis-components}
\small
\setlength{\tabcolsep}{5pt}
\begin{tabular}{rlrrrr}
\toprule
& & \multicolumn{2}{c}{NuMVC} & \multicolumn{2}{c}{MetaVC}\\
\cmidrule(lr){3-4}\cmidrule(lr){5-6}
Vertices & Incidence & Path & All pairs & Path & All pairs\\
\midrule
960 & Quasi-cyclic & $2.67\pm0.33$ & $3.00\pm0.00$ & $2.56\pm0.38$ & $3.00\pm0.00$ \\
960 & Random regular & $2.00\pm0.00$ & $2.22\pm0.19$ & $2.00\pm0.00$ & $2.22\pm0.38$ \\
1,752 & Quasi-cyclic & $6.22\pm0.38$ & $7.11\pm0.19$ & $6.33\pm0.33$ & $6.89\pm0.19$ \\
1,752 & Random regular & $5.89\pm0.19$ & $5.78\pm0.51$ & $5.89\pm0.19$ & $6.11\pm0.51$ \\
\bottomrule
\end{tabular}
\end{table}

The quasi-cyclic design produced larger average shortfalls for both
encodings and solvers at both sizes (Table~\ref{tab:mis-components}). For this design, path encoding
reduced the average shortfall relative to all-pairs encoding in all four
size--solver combinations. For random-regular incidence at 1,752 vertices,
the direction depended on the solver. These comparisons concern the
quality of solutions returned within ten minutes; no setting recovered
an optimum.

\FloatBarrier
\section{Maximum independent set: construction and structural analysis}
\label{app:mis-method}

\subsection{Generation and certificates}

For lift factor $L$, the generator constructs a quasi-cyclic lift of
$K_{3,3}$ with $M=3L$ constraints and variables, following constructions
used in LDPC code design \citep{fossorier2004qc,thorpe2003protographs}.
It tries at most 40 shift configurations, estimating girth by breadth-first
search from up to 30 randomly sampled variable vertices and stopping when this
estimate is at least 12. If that threshold is not reached, it retains the
configuration with the largest estimated girth. It then applies up to
$s\leq3{,}000$ degree-preserving edge swaps, accepting a swap only if
every newly created cycle has length at least 12, with at most $20s$
proposals. The generator records the rank after the swaps.

In the MIS encoding, each constraint
forms a four-vertex clique, and consistency edges connect each variable's
three occurrences along a path. The vertices of the resulting MIS graph
are randomly relabeled. Degrees range from 9 to 15, with equal degrees
within each constraint clique. After the swaps, the incidence matrix
$H\in\FF^{M\times M}$ has rank $M$ for each of the nine benchmark graphs.

The optimality certificate consists of the clique partition and the
planted independent set. The partition gives the upper bound $M$, and
the planted set attains it. Full rank additionally establishes uniqueness
by Proposition~\ref{prop:mis}.

\subsection{Assignment distances}

All matrix operations in this subsection are over $\FF$. For a set $S$ of
violated constraints, let $\mathbf 1_S\in\FF^M$ be its indicator vector. Since
$Hx^*=c$ and $H$ is invertible, the unique assignment with precisely
these violations is
\[
  Hx=c+\mathbf 1_S, \qquad x=x^*+H^{-1}\mathbf 1_S.
\]
Its satisfied constraints select an independent set of size $M-|S|$.
Define
\[
  \beta_t=\min_{|S|=t}\operatorname{wt}(H^{-1}\mathbf 1_S),
\]
where $\operatorname{wt}$ denotes Hamming weight. Thus $\beta_t$ is the
minimum Hamming distance from $x^*$ among assignments violating exactly
$t$ constraints. We computed $H^{-1}$ by Gaussian elimination and
evaluated all columns and all pairs of columns to obtain $\beta_1$,
$\beta_2$ and the mean distance for two violations
(Table~\ref{tab:mis-structure}).

\begin{table}[htbp]
\centering
\caption{Structural quantities and eight-hour solver results for the MIS
family. Optimum is the planted optimum $M$, Best is the largest
independent set found, and Shortfall is $M-\mathrm{Best}$. Patterns gives
$\binom{M}{\mathrm{Shortfall}}$,
rounded to the nearest power of ten: the number of XOR assignments with
that many violated constraints. It does not count arbitrary independent
sets with the same shortfall.}
\label{tab:mis-structure}
\small
\setlength{\tabcolsep}{5pt}
\begin{tabular}{rrrrrrrrr}
\toprule
Vertices & Edges & Optimum & $\beta_2$ & $\beta_2/M$ & Runs & Best & Shortfall & Patterns\\
\midrule
960 & 5{,}280 & 240 & 86 & 0.36 & 24 & 239 & 1 & $10^{2}$ \\
1{,}356 & 7{,}458 & 339 & 122 & 0.36 & 20 & 336 & 3 & $10^{7}$ \\
1{,}752 & 9{,}636 & 438 & 164 & 0.37 & 20 & 433 & 5 & $10^{11}$ \\
2{,}160 & 11{,}880 & 540 & 212 & 0.39 & 15 & 533 & 7 & $10^{15}$ \\
2{,}556 & 14{,}058 & 639 & 256 & 0.40 & 15 & 630 & 9 & $10^{20}$ \\
2{,}952 & 16{,}236 & 738 & 286 & 0.39 & 15 & 729 & 9 & $10^{20}$ \\
3{,}360 & 18{,}480 & 840 & 342 & 0.41 & 15 & 828 & 12 & $10^{26}$ \\
3{,}756 & 20{,}658 & 939 & 392 & 0.42 & 15 & 924 & 15 & $10^{32}$ \\
3{,}996 & 21{,}978 & 999 & 404 & 0.40 & 15 & 984 & 15 & $10^{33}$ \\
\bottomrule
\end{tabular}
\end{table}

\begin{figure}[htbp]
\centering
\includegraphics[width=\linewidth]{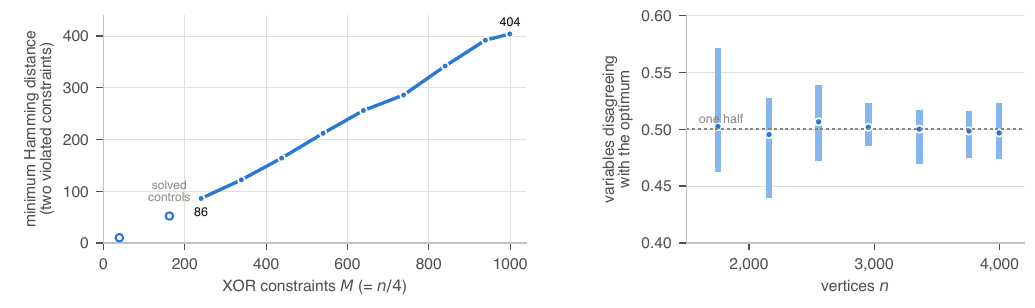}
\caption{Left: exact minimum assignment distance $\beta_2$ for the nine
benchmark graphs (filled markers) and two full-rank solved controls
(open markers). Right: fraction of consistently decoded variables that
disagree with $x^*$ in the available final eight-hour solver sets. Dots
show medians and bars show observed ranges, using 103 of 110 runs on the
seven largest instances.}
\label{fig:mis-structure}
\end{figure}

The observed values of $\beta_2/M$ range from 0.36 to 0.42
(Figure~\ref{fig:mis-structure}, left). For every
graph, the mean distance over all pairs is within one variable of $M/2$.
Invertibility gives $\binom{M}{t}$ assignments with exactly $t$
violations, approximately $10^{33}$ for $M=999$ and $t=15$. Minimum
distances were computed exhaustively only for $t=1,2$.

\subsection{Final solver sets}

A partial independent set need not represent a consistent XOR
assignment: if the middle clique on a variable's path is empty, selected
vertices at the two ends may disagree. For each available final set, we
compared its selected vertices with the planted set and decoded the
variables that receive the same value from every selected vertex
containing them. The median fraction of inconsistent
variables was about 2\% for each graph.

Final sets were available for 103 of the 110 runs on the seven largest
instances. Among consistently decoded variables, the median disagreement
with $x^*$ was approximately one half for every graph
(Figure~\ref{fig:mis-structure}, right); the minimum over
all available sets was about 44\%. Approximately three quarters of
occupied cliques selected a vertex other than the planted one. Thus small
shortfalls coexisted with substantial disagreement with the planted
solution.

\subsection{Recovery of the XOR structure}

The decoder identifies constraint cliques from pair codegrees (numbers
of common neighbors), recovers shared variables from edges between
cliques, and solves the reconstructed XOR system by Gaussian elimination.
Using only the graph file, it returned an optimum for each of the nine
graphs in under 0.08 seconds.

\FloatBarrier
\section{Market Split: additional results}
\label{app:ms-results}

\subsection{Detailed benchmark results}

\begin{figure}[htbp]
\centering
\includegraphics[width=\linewidth]{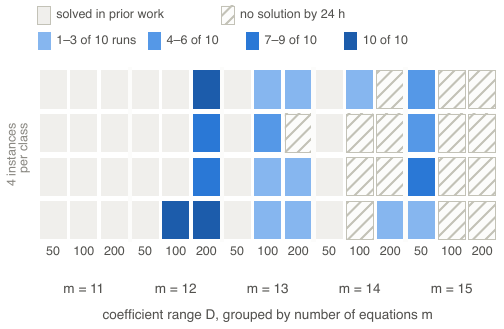}
\caption{QOBLIB Market Split instances with $m\ge11$. Gray indicates
instances solved in prior work; hatching indicates previously open
instances unsolved at the 24-hour cutoff. Blue indicates first reported CPU-only
solutions from this work, shaded by the number of successful runs out
of ten. Each run had two allocated CPU cores.}
\label{fig:frontier}
\end{figure}

\begin{figure}[htbp]
\centering
\includegraphics[width=0.62\linewidth]{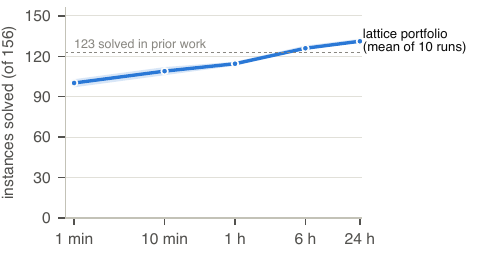}
\caption{Market Split instances solved over time with two allocated CPU
cores per run. For the lattice algorithm, the line shows the mean number
solved over ten seeds and the shaded band shows the observed range.}
\label{fig:coverage}
\end{figure}

Figures~\ref{fig:frontier} and~\ref{fig:coverage} summarize the
instances solved. Table~\ref{tab:first} lists the first reported CPU-only solutions, and
Table~\ref{tab:all} gives results for all 39 QOBLIB classes.

\begin{table}[htbp]
\centering
\caption{The 18 first reported CPU-only solutions in the ten-seed evaluation.
Each run had two allocated CPU cores and a 24-hour cutoff. Success
intervals are pointwise exact 95\% Clopper--Pearson intervals
\protect\citep{clopper1934}, assuming independent trials for each instance.
Times summarize successful runs only; NA denotes an undefined sample
standard deviation because only one run succeeded.}
\label{tab:first}
\small
\setlength{\tabcolsep}{5pt}
\begin{tabular}{llccr}
\toprule
Instance & $(m,n,D)$ & Successes & 95\% interval (\%) & Mean $\pm$ SD (h)\\
\midrule
ms\_12\_100\_003 & $(12,110,100)$ & 10/10 & 69.2--100.0 & $2.86\pm 1.53$ \\
ms\_12\_200\_000 & $(12,110,200)$ & 10/10 & 69.2--100.0 & $7.67\pm 5.84$ \\
ms\_12\_200\_001 & $(12,110,200)$ & 8/10 & 44.4--97.5 & $5.54\pm 6.47$ \\
ms\_12\_200\_002 & $(12,110,200)$ & 9/10 & 55.5--99.7 & $5.70\pm 4.06$ \\
ms\_12\_200\_003 & $(12,110,200)$ & 10/10 & 69.2--100.0 & $4.40\pm 2.72$ \\
ms\_13\_100\_000 & $(13,120,100)$ & 2/10 & 2.5--55.6 & $14.75\pm 11.06$ \\
ms\_13\_100\_001 & $(13,120,100)$ & 5/10 & 18.7--81.3 & $13.27\pm 5.75$ \\
ms\_13\_100\_002 & $(13,120,100)$ & 3/10 & 6.7--65.2 & $17.65\pm 5.96$ \\
ms\_13\_100\_003 & $(13,120,100)$ & 2/10 & 2.5--55.6 & $9.77\pm 2.04$ \\
ms\_13\_200\_000 & $(13,120,200)$ & 1/10 & 0.3--44.5 & $16.61\pm \mathrm{NA}$ \\
ms\_13\_200\_002 & $(13,120,200)$ & 1/10 & 0.3--44.5 & $3.68\pm \mathrm{NA}$ \\
ms\_13\_200\_003 & $(13,120,200)$ & 1/10 & 0.3--44.5 & $13.20\pm \mathrm{NA}$ \\
ms\_14\_100\_000 & $(14,130,100)$ & 1/10 & 0.3--44.5 & $22.85\pm \mathrm{NA}$ \\
ms\_14\_200\_003 & $(14,130,200)$ & 1/10 & 0.3--44.5 & $22.22\pm \mathrm{NA}$ \\
ms\_15\_050\_000 & $(15,140,50)$ & 4/10 & 12.2--73.8 & $11.57\pm 7.37$ \\
ms\_15\_050\_001 & $(15,140,50)$ & 6/10 & 26.2--87.8 & $11.73\pm 8.45$ \\
ms\_15\_050\_002 & $(15,140,50)$ & 7/10 & 34.8--93.3 & $10.77\pm 7.84$ \\
ms\_15\_050\_003 & $(15,140,50)$ & 1/10 & 0.3--44.5 & $8.85\pm \mathrm{NA}$ \\
\bottomrule
\end{tabular}
\end{table}

\begin{table}[p]
\centering
\caption{Results for all 39 QOBLIB classes, each containing four
instances, with a 24-hour cutoff per run. Solved counts are out of
four; lattice-search counts give the range across ten seeds. Times are
means $\pm$ sample standard deviations in seconds over successful runs.
A dash denotes an unavailable statistic; NA denotes a single success.
$\dagger$ marks memory exhaustion on at least one instance
(CPLEX: 16\,GiB host memory; Kempke--Koch: 480\,GiB host or approximately
140\,GiB device memory). ERR denotes an implementation error.}
\label{tab:all}
\footnotesize
\setlength{\tabcolsep}{3pt}
\begin{tabular}{rcrcrcrcr}
\toprule
& \multicolumn{2}{c}{CPLEX}
& \multicolumn{2}{c}{Kempke--Koch}
& \multicolumn{2}{c}{\sd{}}
& \multicolumn{2}{c}{Lattice algorithm}\\
\cmidrule(lr){2-3}\cmidrule(lr){4-5}
\cmidrule(lr){6-7}\cmidrule(lr){8-9}
$n/D$ & Solved & Mean $\pm$ SD
& Solved & Mean $\pm$ SD
& Solved & Mean $\pm$ SD
& Solved & Mean $\pm$ SD\\
\midrule
20/50 & 4 & $0.129\pm 0.0282$ & 4 & $0.578\pm 0.0233$ & 4 & $0.0442\pm 0.0251$ & 4 & $0.382\pm 0.348$ \\
20/100 & 4 & $0.166\pm 0.0247$ & 4 & $0.82\pm 0.342$ & 4 & $0.0305\pm 0.029$ & 4 & $0.302\pm 0.0839$ \\
20/200 & 4 & $0.128\pm 0.0296$ & 4 & $0.711\pm 0.0326$ & 4 & $0.043\pm 0.0249$ & 4 & $0.29\pm 0.0366$ \\
30/50 & 4 & $2.31\pm 0.935$ & 4 & $0.714\pm <0.001$ & 4 & $0.0562\pm 0.00152$ & 4 & $0.329\pm 0.11$ \\
30/100 & 4 & $3.59\pm 1.8$ & 4 & $0.45\pm 0.0257$ & 4 & $0.0559\pm <0.001$ & 4 & $0.318\pm 0.0781$ \\
30/200 & 4 & $4.9\pm 2.4$ & 4 & $0.606\pm 0.15$ & 4 & $0.0557\pm <0.001$ & 4 & $0.327\pm 0.132$ \\
40/50 & 4 & $51.6\pm 64.9$ & 4 & $0.411\pm 0.0584$ & 4 & $0.0558\pm <0.001$ & 4 & $0.373\pm 0.224$ \\
40/100 & 4 & $664\pm 306$ & 4 & $0.499\pm 0.0762$ & 4 & $0.0681\pm 0.0254$ & 4 & $0.325\pm 0.0376$ \\
40/200 & 4 & $215\pm 218$ & 4 & $0.576\pm 0.0484$ & 4 & $0.0681\pm 0.0249$ & 4 & $0.372\pm 0.145$ \\
50/50 & 4 & $2020\pm 1470$ & 4 & $0.565\pm 0.138$ & 4 & $1.52\pm 1.19$ & 4 & $0.399\pm 0.127$ \\
50/100 & 0$^{\dagger}$ & --- & 4 & $0.838\pm 0.173$ & 4 & $2.08\pm 0.818$ & 4 & $0.442\pm 0.164$ \\
50/200 & 0$^{\dagger}$ & --- & 4 & $1.18\pm 0.485$ & 4 & $2.29\pm 1.97$ & 4 & $0.391\pm 0.101$ \\
60/50 & 0$^{\dagger}$ & --- & 4 & $0.876\pm 0.0754$ & 4 & $2.74\pm 1.54$ & 4 & $0.428\pm 0.129$ \\
60/100 & 0$^{\dagger}$ & --- & 4 & $3.47\pm 1.37$ & 4 & $4.54\pm 2.46$ & 4 & $0.731\pm 0.362$ \\
60/200 & 0$^{\dagger}$ & --- & 4 & $4.96\pm 1.81$ & 4 & $5.07\pm 2.11$ & 4 & $0.572\pm 0.246$ \\
70/50 & 0$^{\dagger}$ & --- & 4 & $2.11\pm 0.612$ & 4 & $5.27\pm 2.59$ & 4 & $0.595\pm 0.205$ \\
70/100 & 0$^{\dagger}$ & --- & 4 & $11.6\pm 3.79$ & 4 & $7.91\pm 2.13$ & 4 & $2.3\pm 1.7$ \\
70/200 & 0$^{\dagger}$ & --- & 4 & $32.6\pm 12$ & 4 & $10.2\pm 1.83$ & 4 & $1.78\pm 0.954$ \\
80/50 & 0$^{\dagger}$ & --- & 4 & $16.5\pm 4.72$ & 4 & $3.02\pm 0.637$ & 4 & $1.34\pm 0.763$ \\
80/100 & 0$^{\dagger}$ & --- & 4 & $166\pm 95.5$ & 4 & $15\pm 11$ & 4 & $8.25\pm 3.44$ \\
80/200 & 0$^{\dagger}$ & --- & 4 & $383\pm 64.5$ & 4 & $79.1\pm 63$ & 4 & $6.13\pm 2.99$ \\
90/50 & 0$^{\dagger}$ & --- & 4 & $124\pm 61$ & 4 & $8.32\pm 2.17$ & 4 & $4.95\pm 2.59$ \\
90/100 & 0$^{\dagger}$ & --- & 1 & $64700\pm \mathrm{NA}$ & 4 & $1290\pm 1110$ & 4 & $41.9\pm 33.7$ \\
90/200 & 0$^{\dagger}$ & --- & $\mathrm{ERR}$ & --- & 4 & $15700\pm 16100$ & 4 & $42.7\pm 41.8$ \\
100/50 & 0$^{\dagger}$ & --- & 4 & $30900\pm 7320$ & 4 & $27\pm 19.8$ & 4 & $14\pm 5.5$ \\
100/100 & 0$^{\dagger}$ & --- & 0 & --- & 2 & $12200\pm 5140$ & 4 & $1100\pm 1270$ \\
100/200 & 0$^{\dagger}$ & --- & 0 & --- & 1 & $6050\pm \mathrm{NA}$ & 4 & $3160\pm 3270$ \\
110/50 & 0$^{\dagger}$ & --- & 0$^{\dagger}$ & --- & 4 & $687\pm 472$ & 4 & $120\pm 113$ \\
110/100 & 0$^{\dagger}$ & --- & 0$^{\dagger}$ & --- & 2 & $83200\pm 4400$ & 4 & $13200\pm 10900$ \\
110/200 & 0$^{\dagger}$ & --- & 0$^{\dagger}$ & --- & 0 & --- & 2--4 & $21000\pm 17500$ \\
120/50 & 0$^{\dagger}$ & --- & 0$^{\dagger}$ & --- & 4 & $5870\pm 6900$ & 4 & $1550\pm 1680$ \\
120/100 & 0$^{\dagger}$ & --- & 0$^{\dagger}$ & --- & 0 & --- & 0--3 & $50500\pm 22000$ \\
120/200 & 0$^{\dagger}$ & --- & 0$^{\dagger}$ & --- & 0 & --- & 0--1 & $40200\pm 24100$ \\
130/50 & 0$^{\dagger}$ & --- & 0$^{\dagger}$ & --- & 2 & $37000\pm 44000$ & 4 & $13200\pm 6860$ \\
130/100 & 0$^{\dagger}$ & --- & 0$^{\dagger}$ & --- & 0 & --- & 0--1 & $82300\pm \mathrm{NA}$ \\
130/200 & 0$^{\dagger}$ & --- & 0$^{\dagger}$ & --- & 0 & --- & 0--1 & $80000\pm \mathrm{NA}$ \\
140/50 & 0$^{\dagger}$ & --- & 0$^{\dagger}$ & --- & 0 & --- & 1--3 & $40200\pm 26200$ \\
140/100 & 0$^{\dagger}$ & --- & 0$^{\dagger}$ & --- & 0 & --- & 0 & --- \\
140/200 & 0$^{\dagger}$ & --- & 0$^{\dagger}$ & --- & 0 & --- & 0 & --- \\
\bottomrule
\end{tabular}
\end{table}

\subsection{Component and repeated baseline comparisons}
\label{app:ms-components}

Using the same final solver as in the ten-seed evaluation, we ran each variant with five
different random seeds and five repetitions of \sd{} on each of
the 156 QOBLIB instances. Each run used two cores on an Intel Xeon
Platinum 8468 processor, an 8\,GiB memory reservation and a 24-hour cutoff.
Timing starts at process launch and includes startup, reduction and
enumeration. Each of the two \sd{} processes had a 3\,GiB address-space
limit; the baseline in Table~\ref{tab:frontier} instead used 16\,GiB for
both processes together. This repeated study is separate from the
ten-seed evaluation and the four-algorithm comparison.

The one-basis variant keeps one basis per process, both path costs and
continued reduction. The variant without continued reduction keeps all
24 basis configurations and both path costs, but stops reduction at
block size 40. Each single-cost variant keeps all 24 configurations and
continued reduction. Every variant retains the geometric bound and the
search schedule.

\begin{table}[htbp]
\centering
\caption{Instances solved at 24 hours in five repetitions of the
156-instance suite. Each repetition counts distinct instances solved;
the previously open subset contains 33 instances. Ranges, means and
sample standard deviations are over the five repetitions.}
\label{tab:ms-components}
\small
\setlength{\tabcolsep}{7pt}
\begin{tabular}{lrrrr}
\toprule
& \multicolumn{2}{c}{All 156 instances}
& \multicolumn{2}{c}{Previously open 33}\\
\cmidrule(lr){2-3}\cmidrule(lr){4-5}
Algorithm & Range & Mean $\pm$ SD & Range & Mean $\pm$ SD\\
\midrule
Full algorithm & 130--133 & $131.60\pm 1.52$ & 7--10 & $8.60\pm 1.52$ \\
One basis per process & 126--131 & $128.60\pm 2.30$ & 3--8 & $6.00\pm 2.12$ \\
No continued reduction & 125--129 & $127.20\pm 1.48$ & 3--7 & $5.00\pm 1.58$ \\
Rank cost only & 128--131 & $129.60\pm 1.34$ & 6--8 & $7.20\pm 0.84$ \\
Gaussian cost only & 126--131 & $128.20\pm 1.92$ & 7--10 & $8.20\pm 1.10$ \\
\sd{} & 113--114 & $113.60\pm 0.55$ & 0 & $0\pm 0$ \\
\bottomrule
\end{tabular}
\end{table}

The full algorithm solved the most instances on average. On paired
instance--seed runs, it succeeded where the one-basis variant failed
in 22 cases, versus 7 with the reverse outcome. The corresponding counts
were 29 versus 7 without continued reduction, 13 versus 3 for rank cost
only, and 26 versus 9 for Gaussian cost only.

For each seed, let $d$ be the full algorithm's solved count minus the
variant's count. Table~\ref{tab:ms-uncertainty} reports
$\bar d\pm 2.776\,\mathrm{SD}(d)/\sqrt{5}$, where $\bar d$ and
$\mathrm{SD}(d)$ are the mean and sample standard deviation of $d$ over
the five seeds and 2.776 is the 0.975 quantile of Student's $t$
distribution with four degrees of freedom. Each replicate covers the entire fixed
suite, retaining the pairing between algorithms. These pointwise
intervals assume independent, approximately normal seed-level
differences; they are approximate with five replicates and are not
adjusted for the four comparisons. Table~\ref{tab:matched-times} compares
the runtimes of \sd{} and the lattice algorithm on instances that both
solved in all five runs.

\begin{table}[htbp]
\centering
\caption{Additional instances solved by the full algorithm at 24 hours:
mean paired differences and pointwise 95\% $t$ intervals across five
seeds. Each repetition contains the same 156 instances.}
\label{tab:ms-uncertainty}
\small
\begin{tabular}{lrr}
\toprule
Comparison variant & Mean difference & 95\% interval\\
\midrule
One basis per process & 3.0 & $[-1.2,\,7.2]$ \\
No continued reduction & 4.4 & $[1.2,\,7.6]$ \\
Rank cost only & 2.0 & $[-0.6,\,4.6]$ \\
Gaussian cost only & 3.4 & $[0.8,\,6.0]$ \\
\bottomrule
\end{tabular}
\end{table}

\begin{table}[htbp]
\centering
\caption{Runtime on common solved instances in classes with at least
100 variables. A row includes only instances solved in all five runs
of both algorithms. Each entry is the median across instances of the
per-instance median runtime over repetitions. Runs used the same CPU
model, two cores, process-launch timing and a 24-hour cutoff. Times
below 0.01 hours are shown as $<0.01$.}
\label{tab:matched-times}
\small
\begin{tabular}{rrrrrr}
\toprule
$m$ & $n$ & $D$ & \shortstack{Common\\instances} &
\shortstack{\sd{}\\median (h)} &
\shortstack{Lattice algorithm\\median (h)}\\
\midrule
11 & 100 & 50 & 4 & $<0.01$ & $<0.01$ \\
11 & 100 & 100 & 2 & 3.41 & 0.08 \\
11 & 100 & 200 & 1 & 1.66 & 1.24 \\
12 & 110 & 50 & 4 & 0.15 & 0.03 \\
13 & 120 & 50 & 4 & 1.20 & 0.27 \\
14 & 130 & 50 & 2 & 10.06 & 2.86 \\
\bottomrule
\end{tabular}
\end{table}

\FloatBarrier
\section{Market Split: algorithm details}
\label{app:ms-method}

\paragraph{Basis configurations.}
Each run uses its seed to sample 24 configurations, varying the column
permutation, the number of BKZ tours (two or four), and whether to apply
a unimodular perturbation of strength one to three followed by
re-reduction. Initial reduction uses block sizes 20, 30 and 40 with the
default BKZ 2.0 strategies. Even- and odd-indexed configurations are
assigned to the two processes.

\paragraph{Geometric pruning.}
Enumeration applies \eqref{eq:ms-geometric} from coordinate $n-m$ down
to coordinate 1. Choosing $e_k$ determines the squared residual
$(\sum_{j=k}^{n-m}R_{kj}e_j-y_k)^2$, which is added to the partial sum.
A branch is pruned when that sum exceeds $E_{\max}$; a complete vector
passes the geometric test when its sum is strictly below $E_{\max}$.
Because QR factorization and enumeration use floating-point arithmetic,
the solver reconstructs each such vector in the original variable order
and checks that every entry is 0 or 1 and that the original integer
equations hold before reporting a solution. A vector that fails either check is discarded, and the search
resumes.

\paragraph{Path costs.}
For coefficient $e_k$, given $e_{k+1},\ldots,e_{n-m}$, the conditional
center is
\[
  \hat e_k=\frac{y_k-\sum_{j>k}R_{kj}e_j}{R_{kk}}.
\]
Children are visited in order of distance from $\hat e_k$, beginning with
the nearest integer. The nearest child has rank zero. Under the rank
cost, it costs zero, and a child of rank $\ell\in\{1,2,3\}$ costs
\[
  \log\frac{1-p_k}{p_k w_\ell},
  \qquad (w_1,w_2,w_3)=(0.88,0.10,0.02).
\]
Here $p_k$ is the probability, averaged over a uniform fractional center,
that a Gaussian offset with scale $1/(2|R_{kk}|)$ leaves the
nearest-integer cell.

The Gaussian cost of child $u$ is
\[
  a_k\bigl[(u-\hat e_k)^2-f_k^2\bigr]
  +\log Z'_k(|f_k|)-\log Z'_k(0),
  \qquad
  a_k=2R_{kk}^2,\quad f_k=\hat e_k-\mathrm{nint}(\hat e_k),
\]
where $\mathrm{nint}$ denotes the nearest integer and
\[
  Z'_k(f)=\sum_{j=-40}^{40}
  \exp\!\left[-a_k\bigl((j-f)^2-f^2\bigr)\right].
\]
The correction from this truncated normalizer is tabulated for
$0\le f\le1/2$ and interpolated at the current fractional center.
Under the Gaussian cost, children of ranks zero through eight are
considered. Each basis has one traversal per path cost, with equal node budgets and separate
adaptive thresholds.

\paragraph{Search and reduction schedule.}
Building a configuration creates two searches, one per path cost, with
fresh QR, projection and cost data, a threshold of 0.5, an empty traversal
history and a resumable search state. The basis is then queued for continued
reduction. A failed basis construction is skipped. Each process first
builds configurations until one succeeds or its configurations or time
are exhausted.

In round $r=0,1,\ldots$, each process attempts further configurations up
to a total of $\min(2^r,12)$ and grants each search $2^{16+r}$ nodes.
An interrupted traversal resumes in the next round, whereas a completed
traversal restarts at a higher threshold and uses the remaining nodes
in the current round. The increment is predicted from the last two
completed traversals to multiply the visited-node count by approximately
2.5 and is clipped to $[0.25,3]$. This estimate uses the slope of the
logarithm of node count against the cost threshold; if it is unavailable
or the slope is at most $10^{-6}$, the increment is 0.5.

Before each search, starting 600 seconds after launch, the process
continues reducing its oldest queued basis with block sizes 50, 60 and 70
whenever its cumulative continued-reduction time is less than half the
time elapsed since that mark. It repeats this check while a basis is
queued and time remains. A basis that completes continued reduction
replaces its two searches with fresh ones, including recomputed QR,
projection and cost data. The process retains the existing searches if
continued reduction fails. It
iterates over the searches present at the start of the round, skips any
that have since been replaced, and first visits the new searches in a
subsequent round.

The first verified solution from either process ends the run. The time
budget is checked between search steps and reductions; termination without a solution
does not establish infeasibility.
\section{Research episodes and development}
\label{app:episodes}
\label{app:chronology}

We selected six MIS episodes and five Market Split episodes in which a
proposal led to an implemented experiment and a subsequent decision.
They follow the development of the final generator and solver and
include rejected proposals and revised assessments
(Tables~\ref{tab:mis-episodes} and~\ref{tab:ms-episodes}). Several episodes
concern successive tests of the same proposal; they are stages within
the two projects. One author classified the sources of proposals and
interpreted the episodes.

\begin{table}[htbp]
\centering
\caption{MIS episodes included in the process analysis.}
\label{tab:mis-episodes}
\small
\begin{tabularx}{\linewidth}{@{}>{\raggedright\arraybackslash}p{0.21\linewidth}YY@{}}
\toprule
Proposal or question & Test outcome & Subsequent decision\\
\midrule
Ideator proposes quasi-cyclic lifts and a concatenation of small parity modules &
The tested parity-module instances were solved. &
Worker implemented the quasi-cyclic alternative from the same proposal.\\[3pt]

Worker makes early hardness claims for quasi-cyclic instances &
Repeated seeds and corrected resource allocation overturned several claims. &
Worker withdrew the claims and retained the generator for further testing.\\[3pt]

Ideator proposes keeping local choices ambiguous after a vertex is fixed &
Four instances reached the target ambiguity but were solved within ten minutes. &
Worker rejected the generator built on this ambiguity measure.\\[3pt]

Human requires resistance to both CPLEX and local search &
CPLEX solved the candidate in 42.5 seconds; two of three fresh ReduMIS seeds also solved it. &
Candidate selection required tests with both solvers.\\[3pt]

Ideator proposes replacing some XOR constraints with exactly-one constraints &
CPLEX solved the resulting instances in 0.0--6.9 seconds. &
Worker rejected the mixed-constraint generator.\\[3pt]

Human asks which retained generator is strongest &
Stored comparisons identified quasi-cyclic instances as promising for the revised size target. &
Worker returned to the quasi-cyclic generator and scaled it.\\
\bottomrule
\end{tabularx}
\end{table}

\begin{table}[htbp]
\centering
\caption{Market Split episodes included in the process analysis.}
\label{tab:ms-episodes}
\small
\begin{tabularx}{\linewidth}{@{}>{\raggedright\arraybackslash}p{0.21\linewidth}YY@{}}
\toprule
Proposal or question & Test outcome & Subsequent decision\\
\midrule
Ideator proposes equivalent lattice bases &
A known-solution diagnostic showed large variation in search cost across bases. &
Following ideator feedback, worker tested search and selection using only problem data.\\[3pt]

Worker ranks bases by an entropy score computed before search &
The three solving bases and all candidate bases had similar mean scores, 11.8 and 11.7. &
Worker rejected the ranking rule; human requested a more informative search assessment.\\[3pt]

Human asks how search behavior varies across bases &
Short searches solved the smaller of two instances on eight of 114 bases and the larger on none of 27. &
Worker continued testing node-limited searches and later revised its interpretation of the failure on the larger instance.\\[3pt]

Worker interprets a small comparison of adaptive and random reduction settings as a general limit &
In small samples, adaptive and random choices of reduction settings gave similar best results. &
After further challenge, worker withdrew the limit claim and implemented a stronger solver.\\[3pt]

Worker allocates search by smallest observed violation &
One basis consumed more than five million candidate checks without a solution. &
Worker adopted iterative deepening with equal node budgets for all bases, keeping violation only as a tie-break.\\
\bottomrule
\end{tabularx}
\end{table}

\paragraph{Additional MIS development details.}
Early resource checks found 24 solver runs sharing 16 cores; the affected
experiments were repeated with the intended allocation. After returning
to the quasi-cyclic generator, the worker scaled it, sparsified the
consistency edges and tested girth and degree variants. An auditing
agent found the graph-only decoder described in
Section~\ref{sec:mis-limits} and identified two solver implementations as
reimplementations or modified course code. The human then required
published implementations for the final evaluation.

\paragraph{Additional Market Split development details.}
The initial known-solution diagnostic found a $2^{12}$-fold difference
between the best and median bases at 90 variables. This motivated the
subsequent tests of basis selection using only problem data. Iterative
deepening with equal budgets solved three of four development instances with $n=90$ and
$D=200$. Later, a request to match a two-core budget led the worker to replace a
discrepancy search over a fixed window of coordinates with cost-bounded
enumeration over all coordinates. The human questioned whether stronger
reduction justified its wall-clock cost. Measurements favored stronger
reduction on 14 of 18 hard instances and cheaper reduction on 113 of
127 mostly easy instances. The final configuration interleaves reduction
with enumeration and uses half as many basis configurations as the
earlier solver.

\FloatBarrier
\section{Research corpus and coding procedure}
\label{app:corpus}

The preserved worker histories cover seven research phases
(Table~\ref{tab:corpus-phases}), comprising 264 human-channel records,
2,508 worker tool calls and 184 identified programs. The discovery records
include four ideator messages for MIS and sixteen for Market Split. August
changes to the Market Split solver appear in the repository history but
could not be attributed to a preserved agent session and are excluded
from this analysis.

\begin{table}[htbp]
\centering
\caption{Research phases represented in the preserved worker histories.}
\label{tab:corpus-phases}
\small
\begin{tabular}{lrrl}
\toprule
Phase & Human records & Tool calls & Worker model\\
\midrule
MIS early investigation & 36 & 379 & Opus 4.8\\
MIS discovery and first release & 86 & 634 & Opus 4.8, 5\\
MIS larger instances & 23 & 356 & Opus 5, 5.5\\
Market Split first approach & 15 & 138 & Opus 4.8\\
Market Split discovery & 9 & 84 & Opus 4.8\\
Market Split scaling & 55 & 548 & Opus 4.8\\
Market Split two-core redesign & 40 & 369 & Opus 5.5\\
\bottomrule
\end{tabular}
\end{table}

\subsection{Human-message inventory}
\label{app:human-inventory}

Of the 264 records, 23 relayed agent-written text and 85 were status
checks or bare continuations. One author assigned each of the remaining
156 records a primary code from the fixed scheme in
Table~\ref{tab:human-codes}. Composite records also received secondary
codes for separable human-authored content; the counts and percentages
below use primary codes only.

\begin{table}[htbp]
\centering
\caption{Primary codes for the 156 human-originated records remaining
after excluding relays and status checks.}
\label{tab:human-codes}
\small
\begin{tabularx}{\linewidth}{@{}lYr@{}}
\toprule
Code & Definition & Records\\
\midrule
Objective & Research goals and scope & 19\\
Evidence & Acceptance criteria, validity and fairness checks & 49\\
Conduct & Interpretation of results and research procedure & 14\\
Compute & Computational budgets and run control & 14\\
Mechanism & Human-originated methods or design constraints & 6\\
Selection & Endorsement of an agent proposal & 5\\
Question & Non-directive queries & 6\\
Packaging & Documents and handoffs & 28\\
Infrastructure & Access and session operations & 15\\
\midrule
Total & & 156\\
\bottomrule
\end{tabularx}
\end{table}

Objective, Evidence, Conduct and Compute account for 96 records (62\%);
Packaging and Infrastructure account for 43. Six records (4\%) received
the Mechanism code. These concerned higher girth, an MIS gadget,
spectral analysis, a compiled enumerator, a reduction block-size limit,
and the wall-clock cost of stronger reduction. Message counts describe
the recorded interaction and do not measure intellectual contribution
or effort.

\subsection{Experimental programs}

Of 97 identified MIS programs, four form the final generation package:
the generator and three verifiers. Of 87 Market Split programs, seven
belong to the final solver and evaluation code. The remaining files
include diagnostics, evaluation tools and rejected candidates. These are
file counts; program size and development effort vary.

\FloatBarrier

\end{document}